\documentclass{article}

\usepackage{microtype}
\usepackage{graphicx}
\usepackage[dvipsnames]{xcolor} 
\usepackage{subfigure}
\usepackage{booktabs} 

\usepackage{hyperref}

\usepackage[accepted]{icml2023}

\usepackage{amsmath}
\usepackage{amssymb}
\usepackage{mathtools}
\usepackage{amsthm}

\usepackage[capitalize,noabbrev]{cleveref}

\theoremstyle{plain}

\theoremstyle{definition}

\theoremstyle{remark}

\usepackage[textsize=tiny]{todonotes}

\icmltitlerunning{Towards discovery of the differential equations}

\begin{document}

\twocolumn[
\icmltitle{Towards true discovery of the differential equations}



\icmlsetsymbol{equal}{*}

\begin{icmlauthorlist}
\icmlauthor{Alexander Hvatov}{itmo}
\icmlauthor{Roman Titov}{itmo}
\end{icmlauthorlist}

\icmlaffiliation{itmo}{NSS Lab, ITMO University, Saint-Petersburg, Russia}

\icmlcorrespondingauthor{Alexander Hvatov}{alex\_hvatov@itmo.ru}

\icmlkeywords{Machine Learning, ICML}

\vskip 0.3in
]



\printAffiliationsAndNotice{}  

\begin{abstract}

Differential equation discovery, a machine learning subfield, is used to develop interpretable models, particularly in nature-related applications. By expertly incorporating the general parametric form of the equation of motion and appropriate differential terms, algorithms can autonomously uncover equations from data. This paper explores the prerequisites and tools for independent equation discovery without expert input, eliminating the need for equation form assumptions. We focus on addressing the challenge of assessing the adequacy of discovered equations when the correct equation is unknown, with the aim of providing insights for reliable equation discovery without prior knowledge of the equation form.
\end{abstract}

\section{Introduction}

Starting from SINDy \cite{brunton2016discovering} and PDE-FIND \cite{rudy2017data}, equation discovery has become an appealing machine-learning tool for obtaining interpretable and concise models from physical-related data. Many SINDy add-ons \cite{messenger2021weak, fasel2022ensemble} have been proposed. All of them (except \cite{fasel2022ensemble}, which we discuss later) provide a single "answer" as the optimization result by employing sparse regression on a large differential equation terms library. This uniqueness of the "answer" also holds for all regression-based frameworks such as \cite{long2019pde}. Another drawback of symbolic regression-like methods is the assumption a priori regarding the terms to include in building the equations and their specific form.

However, when the underlying process is unknown, or we do not expect any specific equation to describe the process, a predetermined term set or equation form becomes impractical. Evolution-based frameworks \cite{xu2020dlga, chen2022symbolic, maslyaev2021partial, atkinson2019data} provide a wider search space, allowing different local optima to be obtained across multiple runs. These frameworks are also less demanding to the amount of a priori assumptions. Nevertheless, the challenge remains to determine which equation is the "correct" one.

In this regard, we need to assess how well the given equation describes the data statistically, which includes considering the uncertainty associated with the equation. In \cite{fasel2022ensemble}, a step forward is taken by combining data and pre-described equation library bootstrapping (the term used in the paper, meaning that we take only the random part of the pre-defined library). As a result, even with gradient methods, varying equations can be obtained from run to run. The underlying idea is that the more frequently we obtain the same equation using a random part of available information, the higher the probability that it accurately describes the process. However, the fixed form $u_t = F(x,u_x,u_{xx},...)$ reduces the likelihood of discovering radically new equations, instead resulting in changes primarily in the coefficients. Alternatively, we can gather all the equations obtained from evolutionary optimizations and assess the variability of each term's coefficients and equation forms. 

The assumption that statistically, we get the same equation type bound with the assumption that there is an underlying physical process in data. Within the paper, we will call the underlying process base, meaning that there could be processes with different natures. Most physical processes exhibit several bases or scales. For example, let us consider the Korteweg-de Vries equation. It could exhibit a simple solid matter transition which is a first-order equation, wave properties in the unbound space determined by the second-order equation and at least complete localized wave properties governed by the third-order equation. A residual part could describe noise  (higher order differential terms) with an ad hoc combination that differs from run to run, whereas the base remains the same. In an ideal world, we must be able to extract all bases with minimal noise part using a discovery algorithm.

To understand how the uncertainty in the coefficients or the entire equation translates into the data description and solution, we must be able to solve every equation encountered during the discovery process. However, existing papers primarily focus on restoring known equations rather than discovering new ones. In such cases, known numerical methods with coefficients as parameters are typically employed.

To address the real equation discovery problem, we need to investigate three main research questions:

\begin{itemize}
\item (Question 1) How do we obtain an equation from the data (equation discovery)?
\item (Question 2) How can we ensure we have obtained the correct equation accurately describing the process (discovery robustness)?
\item (Question 3) How can we solve the equation (equation solution)?
\end{itemize}

From a machine learning perspective, the equation discovery problem leads to several related questions that can be summarized as follows: "How can we ensure that the obtained model has the most descriptive ability?" In this paper, we aim to explore the requirements for making equation discovery a more expert-independent machine learning tool, particularly in cases where the equation is unknown. We propose a working pipeline for equation discovery in such scenarios.

The following text is organized as follows. The next three sections discuss the questions above, present existing solutions, and examine their advantages and drawbacks: Section \ref{sec:equation_discovery} focuses on equation discovery, Section \ref{sec:robust_methods} discusses uncertainty estimation, and Section \ref{sec:equation_solution} covers equation solution methods. Section \ref{sec:summary} provides a brief overview of the state-of-the-art in equation discovery. In Section \ref{sec:proposal}, based on the analysis of the current methods, we present a robust equation discovery pipeline, and Section \ref{sec:conclusion} discusses possible future directions in the field.

\section{Current state of equation discovery}
\label{sec:equation_discovery}

This section presents a typical workflow for a differential equation discovery problem. We assume that the differentiation method corresponding to a given data is chosen, and we can work with differentials of the data fields.

From a machine learning perspective, the equation discovery problem is similar to classical machine learning symbolic regression problems. Given a parametrized model $M(\theta)$, data $D$, and a loss function $L(M(\theta), D)$, we aim to solve the following optimization problem (Eq.~\ref{eq:optimization_problem}):

\begin{equation}
\theta^* = \text{arg} \min \limits_\theta L(M(\theta),D)
\label{eq:optimization_problem}
\end{equation}

The data $D$ typically consists of observations in various spatial, temporal, and other parameter spaces. Exploring different parameter configurations (model types in terms of equation discovery) is advantageous in some cases, so the parameter set $|\theta|$ may change. We may also want to determine the boundary conditions along with the primary model, which is often not addressed in the existing literature. As a baseline, we can fix the boundary conditions and incorporate them into the loss function $L(M(\theta), D)$ as a regularization parameter.

\paragraph{Regression methods.}

Most equation discovery algorithms differ primarily in their model representation and optimization methods. It could be a sparse regression \cite{brunton2016discovering}, neural network regression \cite{long2019pde}, computational graph generic evolution \cite{atkinson2019data}, symbolic graph forest \cite{chen2022symbolic}.

The first group of algorithms, exemplified by SINDy \cite{brunton2016discovering}, uses symbolic (sparse) regression with a pre-defined library of terms. In this approach, the model parameters $\theta$ correspond to the numerical coefficients in the expression (Eq.~\ref{eq:sparse_regression_model}):

\begin{equation}
M(\theta) = \sum \limits _{l_i \in \Lambda} \theta_i l_i
\label{eq:sparse_regression_model}
\end{equation}

Here, $\Lambda=\{l_1,l_2,...\}$ represents the library of terms, and $\theta_l \in R$ is the coefficient for each term. Each term is a differential operator or a product of differential operators applied to an input data field. For example, $l=u \frac{\partial{u}}{\partial{x}}$, where $u$ is a discrete data field, represents such a term. A limitation of this approach is that it assumes that the equation can be expressed solely using the terms present in the pre-defined library. Foretelling all required terms for every equation is a challenging task. Consequently, the library needs to be extended to accommodate any non-linearity that might be present in the observed data. Large libraries can be interactively reduced to obtain a concise expression.

Neural network-based approaches, such as PDEnet \cite{long2019pde}, differ from symbolic regression in model representation. The neural network architecture in these approaches changes whenever additional terms are required in the equation. Both the SINDy and PDEnet methods generally consider equations in parabolic form, as shown in Eq.~\ref{eq:SINDy_form}:

\begin{equation}
u_t=F(u,u_x,...)
\label{eq:SINDy_form}
\end{equation}

Neural ODEs \cite{chen2018neural} can also be seen as a particular case of ODE discovery. However, the model takes on a less transparent form. In the case of neural ODEs, the training process allows us to obtain an equation in the form of Eq.~\ref{eq:neural_ODE_form}:

\begin{equation}
\dot{\textbf{u}}=NN(\textbf{u},t)
\label{eq:neural_ODE_form}
\end{equation}

Here, $NN(\textbf{u},t)$ is a non-linear function approximated by a classical neural network. Although less interpretable, neural ODEs can be used differently to extract the underlying physical properties of the data, deviating from the original aim of extracting the physics of the process.

\paragraph{Evolutionary optimization.} 

The second group of methods for equation discovery relies on evolutionary optimization. Defining a specific type of evolutionary optimization or population algorithm in equation discovery is challenging. In \cite{atkinson2019data}, the algorithm can be classified as a genetic algorithm (GA) with automatic differentiation. However, this approach assumes the analytical form of the solution, limiting its applicability. In this section, we describe principles of EPDE \cite{maslyaev2021partial} --  memetic algorithm -- that combines a genetic algorithm (GA) with a gradient algorithm (in our case, sparse regression). Differentiation is performed numerically. The ability of the algorithm to produce different expressions with each run is a necessary trait to assess robustness further.

We move from complete terms to individual differentials, treating them as "building blocks," and allow evolutionary optimization to construct models using these blocks with greater freedom. We refer to these building blocks as "tokens" and group them into parameterized families $F$, where each family contains elements $f_j(p^{(j)}_1,p^{(j)}_2,...,\overline{x}) \in F$. Typically, only differential operators $F_{diff}=\{\frac{\partial^{p_{n+1}} u}{\partial^{p_1}x_1 ... \partial^{p_n}x_n}\}$ are considered. In addition, we may include the inverse coordinate function $F_{inv}=\{\frac{1}{x_i}\}$ and other parameterized discrete fields $F_{func}=\{v(p_1,...,p_k)\}$. The resulting set of tokens is the union of the selected families $F=\bigcup\limits_j F_j$, where $j$ serves as a placeholder for the token family name.

We employ unified evolutionary optimization procedures regardless of the token families chosen for optimization. Specifically, we learn the model in the form of Eq.~\ref{eq:modelRes}:

\begin{equation}
M(C, \textbf{P} ,\overline{x}) = \sum \limits_{i = 1}^{i \le L}c_i * a_{i}(P_i,\overline{x})
\label{eq:modelRes}
\end{equation}

In Eq.~\ref{eq:modelRes}, $C=\{c_1,...c_L\} \in R^L$ represents the constants (term amplitudes), $a_i(P_i,\overline{x})=\prod \limits_{j=1}^{j=N_{tokens}} f_j(p^{(i)}_1,p^{(i)}_2,...,\overline{x})$ denotes the products of tokens $f_j \in F$ selected from the token families $F$, $P_i=\{p^{(i)}_1,p^{(i)}_2,...\}$ represents the parameter set for term $a_i$, and $\textbf{P}$ represents the parameter multi-index. The number of parameters (and the chromosome size) $|\textbf{P}|$ can change during evolutionary optimization. In most cases, only the observational grid points are considered, i.e., $\overline{x} \in X$. However, the data and model grids may be separated.

The model parameters $\textbf{P}$ are determined using evolutionary optimization. The optimization process involves crossover, which exchanges terms between individuals, and mutation, which can involve token or parameter changes. This process is similar to a standard symbolic regression. To complete the formulation of the evolutionary optimization problem, a fitness function needs to be defined.

We incorporate the gradient step in the fitness function calculation step. We use gradient descent to find coefficients $C$ that minimize the discrepancy. Namely, we solve the optimization problem shown in Eq.~\ref{eq:evo_fitness}:

\begin{multline}
       c_{opt}=\text{arg} \min \limits_c \Big|\Big| a_{target}-\sum \limits_{j=1,...,{target}-1}^{j={target}+1,...,N_{terms}} c_j a_j \Big|\Big|_2 +\\
        + \lambda || c ||_1  , \; c=\{ c_1,...,c_{N_{terms}} \}
        \label{eq:evo_fitness}
\end{multline}

In Eq~\ref{eq:evo_fitness} random term is chosen as ''target'' and balanced with the other terms in a given individual. With this problem formulation, we move away from a fixed equation form (Eq.~\ref{eq:SINDy_form}) and allow for arbitrary equations. Additionally, we introduce l1-regularization using the l1-norm $||\cdot||_1$. Terms with coefficients $c_i \in C \;, c_i < \varepsilon $ below a given threshold $\varepsilon$ are filtered as insignificant. 

The proposed single-objective approach enables several advancements compared to classical problem statements. Firstly, it generates various sub-optimal solutions with each algorithm run, and statistical techniques can be applied for robust optimization, as demonstrated below. Secondly, it introduces fewer restrictions, as the equation is not bound to a pre-defined type and library.

\paragraph{Multi-objective step}

We can incorporate a multi-objective step to introduce some control into the discovery process and enhance population diversity. In this step, we define two objectives: quality and complexity. Quality is associated with the discrepancy metric (Eq.~\ref{eq:evo_fitness}), while complexity refers to the number of terms in the equation. It is worth noting that the number of objectives can be increased if desired.

The multi-objective step is performed using $\text{MOEA}\slash\text{DD}$ optimization \cite{li2014evolutionary}. Once the single-objective fitness computation step is finished, the equations with a specific number of terms are sorted, and the top $L$ equations are selected to form non-dominated Pareto levels $L$, as demonstrated in \cite{maslyaev2021multi}.

As a result,  we obtain a set of Pareto non-dominated equations within a pre-defined range of term numbers. Equations with different objective values of complexity can be viewed as different process scales. It is important to note that multi-objective optimization does not provide an answer to the question "What equation best describes the process?" However, we reduce the choice to Pareto frontier equations.

\paragraph{Answer 1: How to choose the discovery algorithm}

Optimization that allows randomly changing terms is the only choice available when dealing with a completely unknown equation. In the equation discovery, it is usually done using an evolutionary algorithm. In such cases, expert knowledge is replaced by optimization time. Alternatively, we can make expert assumptions regarding the equation's form and utilize regression over a pre-defined library. 

In Section~\ref{sec:proposal}, we show that in contrast to the pre-defined form in the sparse regression case, the described evolutionary algorithm requires only the observation field and its derivatives as input and can still reach the correct model.

\section{Uncertainty Estimation}
\label{sec:robust_methods}

Equation discovery is typically approached as a problem with a known answer, where a large-scale process is assumed to be known beforehand. However, extracting the known process from the observed data can be challenging and often requires specialized differentiation methods \cite{zhang2022parsimony}. Developing robust differentiation schemes and algorithms is an integral part of numerical methods that allow for handling observational data effectively.

The noise problem may be mitigated by the robustness of the discovery method by itself. In the case of a pre-defined library, one approach to introducing diversity in the equations is by randomly pruning the library, as demonstrated in \cite{fasel2022ensemble}. When no specific equation solution is available, one can obtain a distribution of the coefficients of the terms and assess them as marginal distributions (independent of each other). However, in this case, we still use a pre-defined large library and estimate the probability or "likelihood" of a given term appearing in the resulting equation model.

As mentioned earlier, evolutionary optimization can produce multiple equation candidates. Therefore, the term co-appearance becomes essential, leading to a joint coefficient distribution analysis. Bayesian networks can be used to construct complex joint distributions, providing insight into these dependencies. The structural learning of Bayesian networks is out of the paper's score. We refer the reader to the survey \cite{kitson2023survey}. The example of a Bayesian network for the differential equation is shown in Appendix~\ref{app:BN}. In the case of pre-defined library bootstrapping, Bayesian networks can also be utilized to determine how the absence of one term affects the resulting equation, helping to identify the base terms.

\paragraph{Answer 2: How to assess uncertainty}

Joint distributions benefit sparse regression with a pre-defined library and evolutionary discovery algorithms. It is crucial to evaluate the uncertainty of individual terms and their collective occurrence to accurately determine if the discovered equation is indeed foundational. 

Bayesian networks allow one to obtain a joint distribution automatically, as shown in Section~\ref{sec:proposal}.

\begin{figure*}[h!]
\centering
\includegraphics[width=1\linewidth]{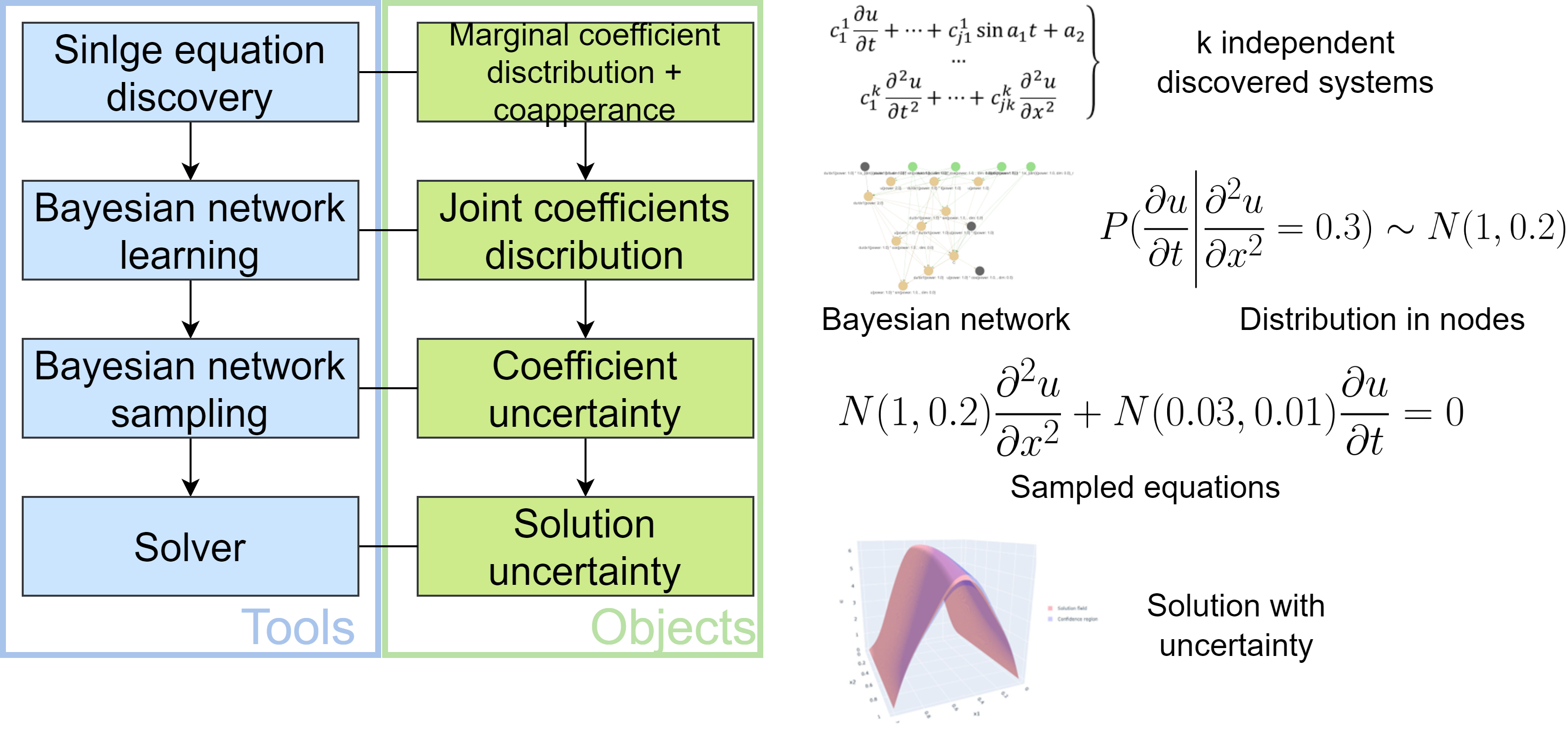}
\caption{Robust equation discovery pipeline: left - a group of tools used in the equation discovery, right - output of the tools with illustrations.}
\label{fig:robust_scheme}
\end{figure*}

\section{Equation Solution}
\label{sec:equation_solution}

In the context of equation discovery, the focus is often on the problem of equation restoration, where we already know the equation and aim to generate data from it to compare it with the discovered equation. Restoring the known equation is a primary quality metric in the literature \cite{brunton2016discovering,chen2018neural,atkinson2019data,chen2022symbolic,fasel2022ensemble}. The approach to equation solution differs for pre-defined library and evolutionary discovery methods.

 In the case of a pre-defined library, as shown in \cite{fasel2022ensemble}, even when bootstrapping the data and library, we obtain a close equation that may only differ in coefficients. This simplifies the equation solution and subsequent data comparison, as we can set up a specific solver for a particular type of equation and vary only the coefficients. Any noise terms are effectively "truncated" in this scenario. Unfortunately, this approach is not applicable in general discovery cases where the equation is unknown.

One possible approach is to unify solutions using physics-informed neural networks (PINNs) such as PINN \cite{raissi2019physics}, and DeepONet \cite{lu2021learning}, along with their various modifications. Typically, these approaches combine data and differential operators within a single loss function, which can introduce potential biases. We require dedicated solvers instead of relying solely on PINNs to mitigate this.

When considering solvers, the choices available for solving multiple classes of equations without extensive tuning are limited. There are a few viable options to consider: a decision tree-based solver written in the Julia programming language \cite{rackauckas2020universal}, a DeepXDE solver based on neural networks in Python \cite{lu2021deepxde} and TEDEouS solver also based on neural networks \cite{hvatov2023automated}. Both of these solvers can be used effectively for equation discovery.

\paragraph{Answer 3: How to solve discovered equations}

The use of more versatile solvers would benefit the equation discovery process. It is not always feasible to "truncate" equations to a known form, nor is it always possible to set up a solver specifically for a single equation type. Therefore, employing solvers that can handle multiple equation types without extensive customization is advantageous.

\section{Equation discovery state-of-the-art}
\label{sec:summary}

In summary, the existing equation discovery methods rely on a pre-defined library and a preset equation form, often assuming a specific type of equation, such as a parabolic equation. These methods assume that the underlying process in the data can be represented using the terms in the library and the chosen equation form. If new terms need to be added to the library, the model training process must be repeated from the beginning. Assessing the completeness of the library requires problem formulation and expert knowledge. Currently, the process of equation discovery relies on the expertise of an area specialist to select equations that may be relevant to the specific discovery task. Furthermore, the evaluation of discovery algorithms often focuses on their ability to restore known equation coefficients rather than discovering equations themselves.

It is necessary to incorporate uncertainty estimation into the equation domain to address unknown equations. Joint distribution analysis becomes important as multiple potential physical bases may exist. A solution algorithm must also be incorporated to transfer equation uncertainty to the data domain.

\section{Proposed approach for expert-independent equation discovery}
\label{sec:proposal}

This section provides an overview of the advancements made towards achieving expert-independent equation discovery. Specifically, we introduce the framework of evolutionary equation discovery, its combination with an equation solver, and use Bayesian networks for robust discovery. The complete pipeline for robust equation discovery is depicted in Figure~\ref{fig:robust_scheme}.

Preliminary experiments demonstrate that combining equation discovery \footnote{\url{https://github.com/ITMO-NSS-team/EPDE}}\cite{maslyaev2021multi} with a solver \footnote{\url{https://github.com/ITMO-NSS-team/torch_DE_solver}} \cite{hvatov2023automated} enables the estimation of solution uncertainties rather than simply coefficients. However, assessing the co-appearance of terms still needs to be improved. Bayesian network learning \footnote{\url{https://github.com/aimclub/BAMT}} \cite{deeva2023advanced} is employed to derive joint distributions from multiple runs of the equation discovery algorithm. An automated tool for Bayesian network construction is utilized to automatically form the joint distribution based on a set of equations obtained within the discovery process.

Subsequently, the equations are sampled from the Bayesian network, starting from an arbitrary left-hand side with a coefficient of one. Each sampled equation is then solved to obtain uncertainties in the solution. The Bayesian network serves as the robust "meta-equation," although individual equation samples can be easily obtained from it.

This combined approach enables automated assessment of uncertainties in both equation coefficients and solution by utilizing joint distributions instead of marginal distributions. Consequently, it becomes possible to statistically analyze how well the discovered equations describe the provided data. Instead of relying on the completeness of the library and a fixed equation form, we focus on evaluating the co-appearance of simple differential terms obtained by the stochastic optimization process.

While each step, including differentiation, optimization, distribution discretization, Bayesian network optimization, and solution, introduces some error, the resulting confidence interval allows for an accurate evaluation of the chosen discovery method's quality.

\subsection{Experimental study}

To illustrate the proposed approach, we revisit the Lynx-Hare dataset \cite{langton1921conservation} that was previously used in \cite{fasel2022ensemble} to restore the Lotka-Volterra model. We use it since it is simple and thus illustrative. We note, however, that approach was tested in other cases, such as Burger's equation and Korteweg -- de Vries equation.

\paragraph{Existing approach.} The Lotka-Volterra model in \cite{fasel2022ensemble} for these data \footnote{Openly available in the authors' repository \url{https://github.com/urban-fasel/EnsembleSINDy/blob/main/SINDY/main_runLotkaVolterra.m}} was obtained using the fixed set of terms for each equation. Namely, the model Eq.~\ref{eq:LVSINDy_params} was used.

\begin{equation}
  \left\{\begin{array}{@{}l@{}}
    \dot{u}= C_1u +C_2uv+C_3v+C_4u^2+C_5v^2+C_6+\\
    +C_7uv^2+C_8vu^2+C_9u^3+C_{10}v^3\\
    \dot{v}=D_1v +D_2uv+D_3u+D_4u^2+D_5v^2+D_6+\\
    +D_7uv^2+D_8vu^2+D_9u^3+D_{10}v^3
  \end{array}\right. 
  \label{eq:LVSINDy_params}
\end{equation}

The optimization uses reduced models, Eq.~\ref{eq:LVSINDy_params}, obtained by randomly canceling terms. Such a process is called library bootstrapping in the paper. The uncertainty of terms was obtained using 1000 bootstrapped models. The resulting marginal coefficient distribution is Eq.~\ref{eq:LVSINDy_res}. It should be noted that $\epsilon=O(10^{-5})$ and the coefficients are represented by the mean values along with two-sided confidence intervals constructed using the 95th percentile of the standard distribution. 

\begin{equation}
  \left\{\begin{array}{@{}l@{}}
    \dot{u}= \color{OliveGreen}(0.5274 \pm  0.0049)u +(-0.025 \pm  \epsilon)vu \\
    \dot{v}= \color{OliveGreen}(-0.9691 \pm  0.1779)v+(0.027 \pm  \epsilon)vu+\\
    +(-0.1193 \pm  0.1047)u+(0.1755 \pm  0.2248)
  \end{array}\right. 
  \label{eq:LVSINDy_res}
\end{equation}

We note that unlike in \cite{fasel2022ensemble}, we do not normalize the data on dispersion. Thus, the coefficients of the terms $uv$ must be multiplied by $16.656$ and $21.414$  for the first and second equations to reach correspondence. The equations in the original paper were solved using the Runge-Kutta method with \texttt{ode45} Matlab solver using the fixed form of equation Eq.~\ref{eq:LVSINDy_params}.

To sum up, to discover the Lotka-Volterra model from data using the sparse regression group of methods, we must know the ``at the shoreline'' equation form, setup solver, and possibly tune the bootstrapping method to get different results from run to run.

As advantages, we get similar to the theoretically obtained coefficients \cite{howard2009modeling} that are shown in Eq.~\ref{eq:LV_theory} (other coefficients are equal to zero).

\begin{equation}
  \left\{\begin{array}{@{}l@{}}
    \dot{u}=0.55u-0.028uv \\ 
    \dot{v}=-0.84v+0.026uv
  \end{array}\right. 
    \label{eq:LV_theory}
\end{equation}

The second advantage is the optimization speed. Methods based on a sparse regression over a pre-defined library work faster than any evolutionary algorithm due to the gradient descent used. Furthermore, modern tuned solvers are also known as fast tools.

\paragraph{Proposed approach.} We employ a multi-objective evolutionary discovery algorithm \cite{maslyaev2021multi} and Bayesian network structural learning \cite{deeva2023advanced} along with a neural network solver \cite{hvatov2023automated} to solve the obtained system. We consider two cases for system discovery: (a) a system bounded with first-order equations and (b) bounded with second-order equations. Unlike pre-defined library algorithms, each equation is treated independently, meaning that the system consists of several interconnected equations rather than a vector representation.

Next, we treat each term in the equation as a random variable and use the entire equation to determine the co-appearance of terms. Based on this information, a Bayesian network represents the joint distributions of term coefficients as shown in Appendix~\ref{app:BN}.

Following Bayesian network analysis, we may identify one or multiple graph components. In our case, there was a single component. We then choose an arbitrary initial node, $\dot{u}$ and $\dot{v}$ for the first and second equations, respectively. Once the initial node is selected, we sample 30 equations to determine the uncertainty in the coefficients and solve them to determine the uncertainty in the data domain.

For case (a), we identify two system bases presented in Eq.~\ref{eq:LVa}. For illustration, all coefficients are shown in a marginal distribution way. However, we note that co-appearance (and, more general, joint distributions) is taken into account. The resulting distribution is a Bayesian network (an example is shown in Appendix \ref{app:BN}). 

\begin{equation}
\begin{array}{cc}
  \left\{\begin{array}{@{}l@{}}
    \dot{u}= \color{OliveGreen}{(0.5598 \pm  \epsilon)u}+(-0.028 \pm  \epsilon)uv \color{black}+\\+(0.0941 \pm  0.1157)\dot{v} +(0.0019 \pm  0.0001)\dot{u}v +\\+(0.0023 \pm  0.0001)\dot{u}\dot{v} -0.1073 \pm  0.008 \\
    \dot{v}=\color{OliveGreen}(-0.8278 \pm  \epsilon)v +(0.0256 \pm  \epsilon) uv \color{black}+\\ +(0.0037 \pm  0.0005)\dot{u}+(-0.0021 \pm  0.0001)\dot{u}\dot{v}+\\ +(0.0998 \pm  0.0045)
  \end{array}\right. \\
  \left\{\begin{array}{@{}l@{}}
    \dot{u}= (18.1103 \pm  \epsilon)  + (-0.5132 \pm  0.0084)v+\\+(0.0263 \pm  0.0017)\dot{v} +(0.0108 \pm  0.0001)u\dot{u}\\
    \dot{v}=(0.8540 \pm  9.1659)+(0.0279 \pm  \epsilon)v\dot{v}
  \end{array}\right.
\end{array}
  \label{eq:LVa}
\end{equation}

The first system in Eq.~\ref{eq:LVa} represents the Lotka-Volterra system, which is essentially correct and marked in green, with additional noise terms determining the variance. The second equation represents a higher-scale process, specifically $\{\dot{u}, \dot{v}\}=\{C_1,C_2\}$, i.e., locally constant population speed increase or decrease.

\begin{figure}[h!]
    \centering
    \includegraphics[width=\linewidth]{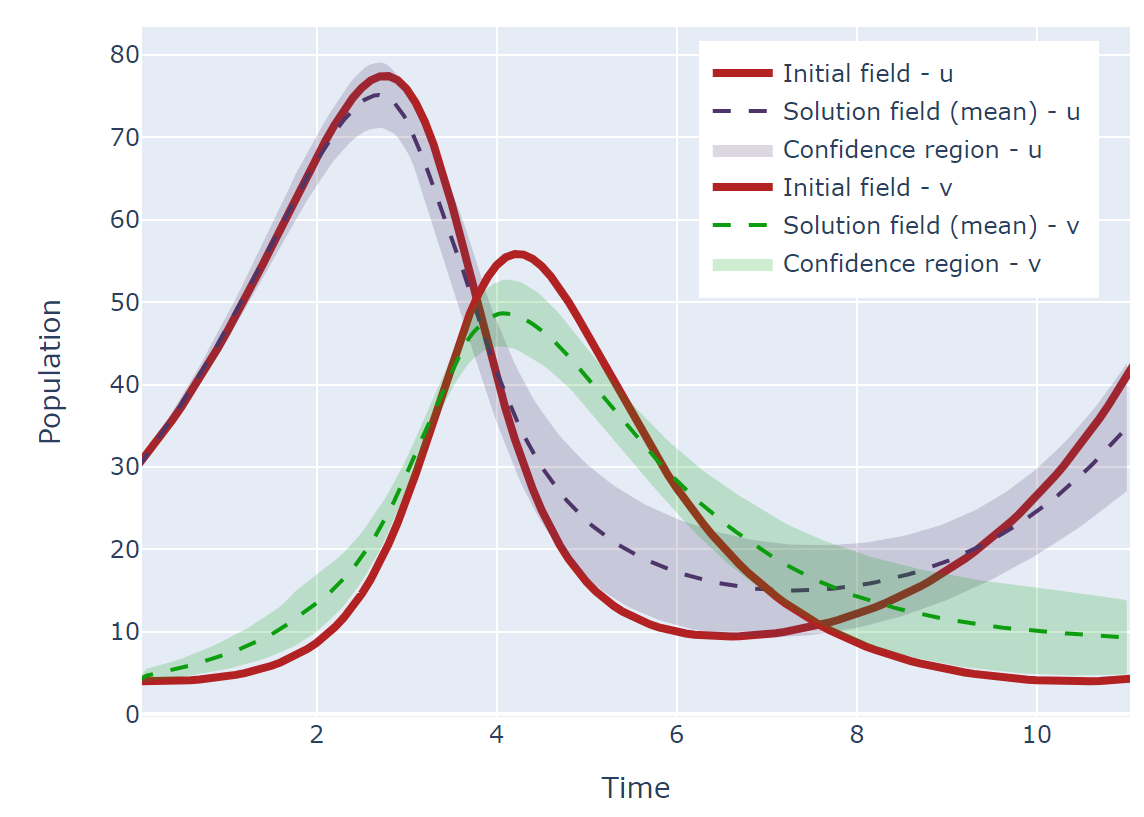}
    \caption{The results of solution sampled using Bayesian network systems for case (a) - first-order derivative restriction. Red lines - data, dashed line - 'mean' solution, corresponding intervals - maximum and minimum of an integrated function at a given time step. }
    \label{fig:LVa}
\end{figure}

In Fig.~\ref{fig:LVa}, the solutions of the sampled equations using the neural network solver are depicted.

We note that for the first system in Eq.~\ref{eq:LVa}, we get almost perfect theoretical coefficients Eq.~\ref{eq:LV_theory}. Moreover, they are closer to E-SINDy Eq.~\ref{eq:LVSINDy_res}. Moreover, the only information used is discrete observations $u$ and $v$ for the Lynx and Hare populations, respectively, and their numerical derivatives $\dot{u}$ and $\dot{v}$.

In case (b), as the complexity increases with second-order terms, we still mainly obtain first-order equations. However, the variability also increases. We can still find bases from case (a) (Eq.\ref{eq:LVa}), as shown in Eq.\ref{eq:LVb}. In particular, the coefficients of the Lotka-Volterra model are close to each other.

\begin{equation}
\begin{array}{cc}
  \left\{\begin{array}{@{}l@{}}
    \dot{u}= \color{OliveGreen}( 0.5426 \pm  \epsilon)u+(-0.0277 \pm  \epsilon)uv+\\+(-0.1194 \pm  0.0009)\dot{v}+(-0.1461 \pm  0.0087)\\
    \dot{v}=\color{OliveGreen}(-0.8535 \pm  \epsilon)v+(0.0260 \pm  \epsilon)uv+\\+(0.1676 \pm  0.1113)u+(1.1087 \pm  0.3570)
  \end{array}\right. \\
  \left\{\begin{array}{@{}l@{}}
    \dot{u}= (17.7157 \pm  0.0008)+(-0.0406 \pm  0.0005)\dot{v}+\\+(-0.7592 \pm  0.0096)v  \\
    \dot{v}=(0.0131 \pm  0.6421)+(-0.8538 \pm  \epsilon)v+\\ +(0.0260 \pm  \epsilon)uv 
  \end{array}\right.
\end{array}
  \label{eq:LVb}
\end{equation}

\begin{figure}[h!]
    \centering
    \includegraphics[width=\linewidth]{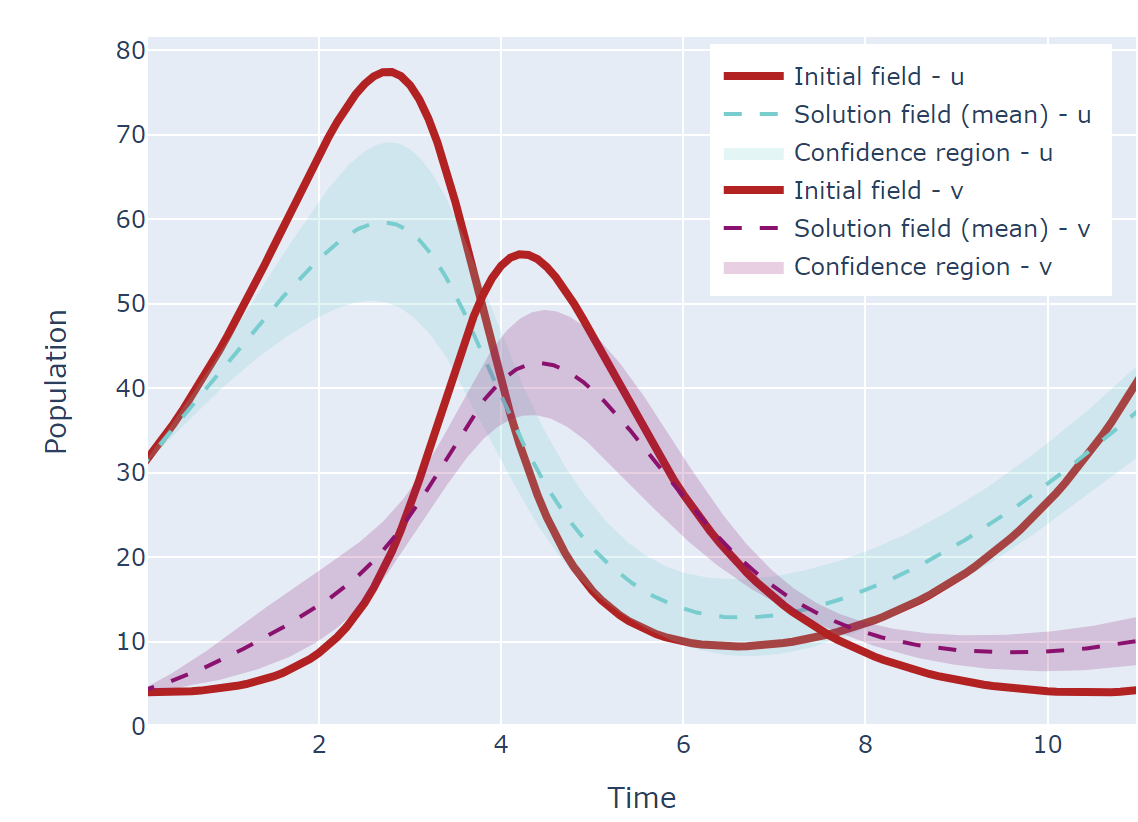}
    \caption{The results of solution sampled using Bayesian network systems for case (b) - second-order derivative restriction. Red lines - data, dashed line - 'mean' solution, corresponding intervals - maximum and minimum of an integrated function at a given time step. }
    \label{fig:LVb}
\end{figure}

The solution of sampled equations for case (b) is shown in Fig.~\ref{fig:LVb}.

Compared to Eq.~\ref{eq:LVa}, the coefficients Eq.~\ref{eq:LVb} are less similar to the theoretical ones Eq.~\ref{eq:LV_theory} but more similar to those obtained with E-SINDy (Eq.~\ref{eq:LVSINDy_res}). However, we can still extract the physical base and fit them more precisely if needed.

We gather the percentage error for all models provided with respect to those obtained in \cite{howard2009modeling} in Tab.~\ref{tab:comparison}. However, since this is not a synthetic case, we do not have a ``right answer'', and the error metrics show nothing.

\begin{table}[h!]
\caption{Coefficient errors with respect to the Eq.~\ref{eq:LV_theory}. The mean error is taken since there is a $uv$ term in both equations.}
\begin{tabular}{|l|l|l|l|l|}
\hline
Terms                              & u    & v    & mean(uv) & mean(error) \\ \hline
E-SINDy                            & 4 \% & 13\% & 7 \%     & 8 \%        \\ \hline
Case (a)  & 2\%  & 1\%  & 1\%      & 1\%         \\ \hline
Case (b) & 1 \% & 2\%  & 1\%      & 1\%         \\ \hline
\end{tabular}
\label{tab:comparison}
\end{table}

The proposed method allows us to discover equations without relying on a fixed structure or pre-defined form. In the Lotka-Volterra case, the input data is $u$ and $v$ as Lynx and Hare population and their numeral differentials up to a given order. The order and the term power are the only restrictions for the algorithm that could easily be extended, as shown with the transition from case (a) to case (b). 

The provided experiment represents a significant extension of the classical pre-defined library pipeline. With this approach, we gain the ability to discover equations in an expert-independent manner. We emphasize that the expert's time is exchanged significantly for running time. For the classical algorithm \cite{fasel2022ensemble}, we talk about minutes, whereas for the proposed one, we talk about hours in complex cases. Nevertheless, rather than simply restoring known equations, we can uncover new ones using a flexible and adaptive framework.

\section{Conclusion and Discussion}
\label{sec:conclusion}

In conclusion, this paper has explored the potential enhancements that can be made to equation discovery to transform it into a more effective machine learning tool. While the existing methods discussed in the paper are successful in their respective tasks, there are several key areas where improvements can be made to enrich the field of equation discovery:

\begin{itemize}
\item The pre-defined library can be redesigned to utilize simpler building blocks, allowing for adding new terms and accommodating different problem types.
\item Single-objective optimization can be replaced with multi-objective optimization techniques to enhance control over the discovery process and enable the exploration of equation systems.
\item Uncertainty estimation must consider joint distributions to provide a more comprehensive understanding of the relationships between terms.
\item Equation solvers can be leveraged to map coefficient uncertainty to the data-solution domain, bridging the gap between discovered equations and their practical applications.
\end{itemize}

By incorporating these proposed changes, equation discovery can evolve into a powerful tool that is not only applicable to known processes and equations but also suitable for partially known processes (e.g., known Navier-Stokes equation and unknown state equation) or even complete ``black-box" scenarios. This aligns with machine learning principles, enabling the extraction of new physical laws from data and empowering computers to creatively propose new equations for researchers to investigate.

The huge disadvantage is the general extended working time. That is the automatization fee. There is still much work to make the proposed approach fully autonomous. As an example, it is required to work with the Pareto frontier automatically, choose the solver hyperparameters, and differentiate the data for the discovery process.

Advancements in all described fields have the potential to revolutionize the field of equation discovery, allowing for more automated and expert-independent approaches that uncover novel insights and contribute to scientific advancements.


\section{Data and code availability}
\label{sec:data}

The results were obtained using openly available data \cite{langton1921conservation}, that is, data set \url{https://github.com/stan-dev/example-models/blob/master/knitr/lotka-volterra/hudson-bay-lynx-hare.csv} was used. The code to reproduce the experiments is available at the repository \url{https://github.com/ITMO-NSS-team/KLR2023_paper}.



\bibliography{references}
\bibliographystyle{icml2023}

\newpage
\appendix
\onecolumn
\section{Bayesian network example.}
\label{app:BN}

\begin{figure}[h!]
    \centering
    \includegraphics[width=\linewidth]{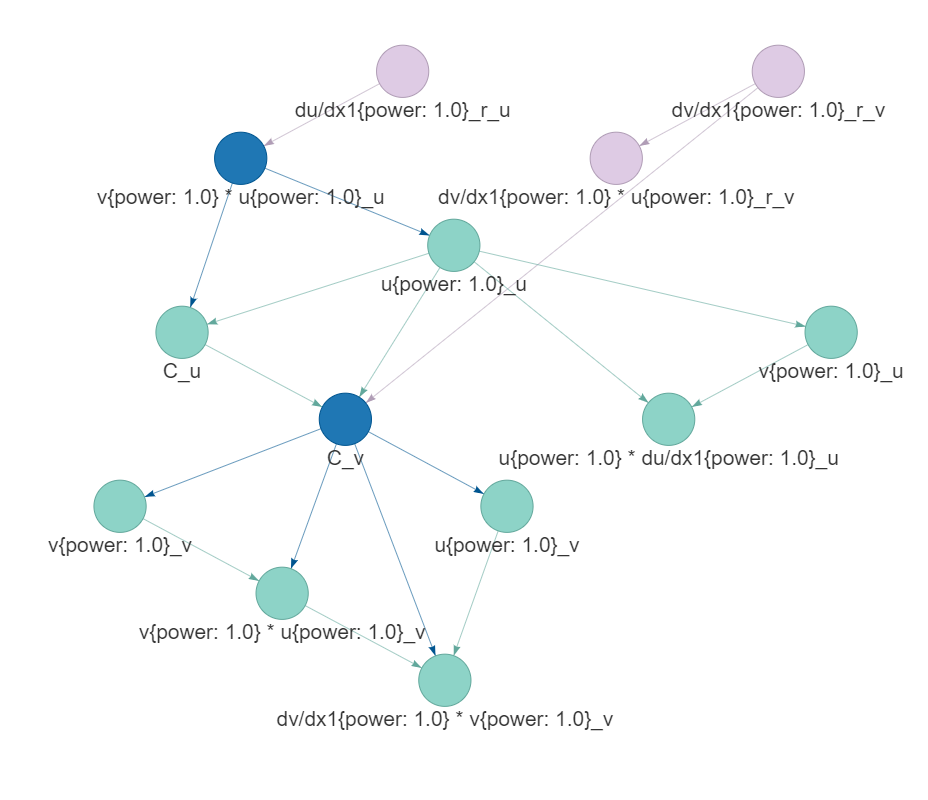}
    \caption{Example of Bayesian network for case (b). Index \text{\_u} determines first equation, index \text{\_v} determines the second.}
    \label{fig:BN_example}
\end{figure}


\end{document}